# TOWARDS AN INTEGRATED VISUALIZATION OF SEMANTICALLY ENRICHED 3D CITY MODELS: AN ONTOLOGY OF 3D VISUALIZATION TECHNIQUES

*Claudine Métral, Nizar Ghoula and Gilles Falquet*

*Centre universitaire d'informatique, University of Geneva, Switzerland*
*claudine.metral@unige.ch*

**ABSTRACT**
3D city models - which represent in 3 dimensions the geometric elements of a city - are increasingly used for an intended wide range of applications. Such uses are made possible by using semantically enriched 3D city models and by presenting such enriched 3D city models in a way that allows decision-making processes to be carried out from the best choices among sets of objectives, and across issues and scales. In order to help in such a decision-making process we have defined a framework to find the best visualization technique(s) for a set of potentially heterogeneous data that have to be visualized within the same 3D city model, in order to perform a given task in a specific context. We have chosen an ontology-based approach. This approach and the specification and use of the resulting ontology of 3D visualization techniques are described in this paper.

## 1. INTRODUCTION

Since the 1960s, the term urban model has been usually related to simplifications and abstractions of real cities, in contrast to its earlier usage referring to ideal cities (Foot, 1981). Today, accurate models can be used to perform, for example, urban simulations (Waddell et al, 2008), building energy consumption (Jones et al, 2000), water quality calculation (Kianirad et al, 2006) or air quality estimation (Moussiopoulos et al 2006). Urban models are widely used by urban planners and designers to explore the city or to plan it prior to acting on it. By this way, urban models can be seen as decision-making tools.

3D city models are specific urban models related to the field of Geographic Information Systems (GIS). 3D city models represent geo-referenced data in three dimensions and under a common view, the virtual city. CityGML (OGC 2008) is the first standard related to 3D city models. CityGML attempts to provide a description of 3D elements like relief, buildings, traffic infrastructure, water bodies, vegetation or city furnitures with their geometry, topology (spatial properties), semantic properties and relevant attributes. The attributes cover classification, function and actual usage of an object. Those objects have been identified to be either required or important in many different application areas. CityGML is also extendable to adapt to the requirements of specific applications. Such extensions have been defined for different issues such as noise, with noise immissions added to buildings, roads or railways (OGC 2008), and for different purposes such as flood information service (Schulte & Coors 2008). We thus obtain enriched 3D city models containing more than mere geometrical elements. In the case described by Schulte and Coors, the basic water model of CityGML has been extended with hydro-numerical data sets, as preliminary step of integration in a Web 3D Service for dynamic 3D flood visualization in interactive 3D scenes. If visualization is effectively one of the main uses of enriched 3D city models it is not the only one. The EU COST Action TU0801 "Semantic enrichment of 3D city models for sustainable urban development" considers 3D city models, semantically enriched with the knowledge underlying in data and models related to urban issues, as potential planning and decision-making tools whose assessment has still to be processed (TU0801 2008).

When designing a 3D virtual city that displays abstract information, the designer faces the problem of choosing the most relevant visualization techniques for viewing abstract information. This means that the selected techniques must at least:
- be able to display the desired information (effectivity);
- efficiently support the user task(s);
- not negatively interfere with each other (e.g. by hiding information).

In this paper we address the problem of integrating visualization techniques for 3D enriched city models by:
- proposing a formal representation of existing visualization techniques, in the form of an ontology of 3D visualization techniques that can be used for computational reasoning;
- showing how selection tasks, such as checking if a technique is compatible with a dataset to display, can be expressed in terms of ontological tasks such as subsumption computation or logical query evaluation.

## 1.1 Visualization techniques for data in 3D Virtual Environments

Enriched 3D city models can be seen as 3D virtual environments (3DVEs) containing not only geometrical elements but also abstract information. Abstract information means all data that cannot be perceptible without a visual abstraction brought into the view of the user. Visualization techniques for data in 3D virtual environments have been devised along with the development of techniques for 3D virtual environments. They are extensively used in simulation environments, video games, tutoring systems, etc. Nevertheless only a few authors have attempted to classify these techniques and to compare and evaluate them.

There are many different approaches in which abstract information can be added to 3D scenes in order to help the user perform specific tasks. Elements to take into consideration for describing these methods include:
- the type of data to visualize (Mackinlay 1986) (Shneiderman 1996);
- the spatial configuration of the 3DVE that can be defined according to the user's viewpoint reference frame (Tyndiuk, 2005), the user being either inside or outside the 3DVE.

Chi (2000) have defined a taxonomy of visualization techniques using a model that differentiates:
- the raw data, called value,
- the meta-data, called analytical abstraction,
- the information that is visualizable on the screen using a visualization technique, called the visualization abstraction, and
- the end-product of the visualization mapping where the user sees and interprets the picture presented to him/her, called the view.

## 1.2 Visualization techniques for enriched 3D city models

Although enriched 3D city models can be considered as 3DVEs, they have some specificities. In fact the use of enriched 3D city models for helping decision making in the urban domain is related to tasks performed by the user of such models and for which he/she has to navigate in or over the 3D model. Typical tasks include evaluation of urban projects in terms of quality of life (including visual aspects), evaluation of the impact of projects on the urban landscape and on other factors.

Such tasks imply the visualization of data that:
- originate from different fields like transport, construction, etc.;
- are of different kinds such as quantitative measures of noise, qualitative soundscapes, …;
- take different forms (from structured data provided by geographical information systems to unstructured documents);
- have different scales (city as a whole, buildings, …);
- are not directly georeferenced (legal text for example although they have a spatiotemporal coverage).

Hence, visualization techniques face issues such as:
- heterogeneity of data implies more possibilities of associated visualization techniques;
- a correlation between the user's context and task requires the usage of a specific visualization technique;
- simple addition/superposition of visualization techniques associated to different data may imply cognitive overload thus generating an incomprehensible 3D scene (too much data, impossible to visualize them together);
- visualization techniques that are suitable for each data taken individually may be incompatible when taken together;

Thus data have to be visualized as a whole from a set of suitable visualization techniques taking in account compatibility - incompatibility rules between the techniques and/or between the data, leading to an integrated visualization.

Given the vast amount of visualization techniques that have be developed over the last decades, and the variety of visual contexts, user tasks, and data types, the selection of relevant visualization techniques is far from trivial. For some techniques work has been performed to evaluate to which task(s) and to which context(s) they are relevant. When the evaluation results are publicly available they can be used to help selecting a technique. An example of such a case is (Vaaraniemi et al 2012) who propose two new visualization techniques for enhancing the visibility of road labels in 3D navigation maps, the expected enhancement being defined relatively to the baseline approach (which is used in almost all existing navigation systems such as Google Earth) and confirmed by a user study. In fact we usually don't have such evaluation results. On the contrary many visualization techniques are used in 3D geographic information systems (3D GIS) by many users and for many tasks without any evaluation. While not formally assessed the utility of such visualization techniques still exists. We also have another case related to 3D video games such as Simcity, for example, which is a city-building and urban planning simulation computer game (SimCity

1989). In the case of video games if intensive evaluations of the 3D visualization techniques used have been performed, their results are usually kept secret.

**1.3 Visualization ontologies**

Different classifications, terminologies, taxonomies or ontologies have been defined in the field of visualization, with different aims. Shu et al (2007) present the design of a visualization ontology, which aims at providing more semantics for the discovery of visualization services. The Top Level Visualization Ontology (TLVO) defined by Brodlie et al (2004) aims at providing a common vocabulary to describe visualization data, processes and products. Based on an analysis of visualization taxonomies and on recent work in visualization ontologies, Morell Pérez et al (2011) propose some modifications to the TLVO in order to better represent the visualization process and data models. Voigt & Polowinski (2011) aim at developing a unifying ontology, which is applicable in visualization systems. By systematically reviewing existing models and classifications, they found in particular that most visualization knowledge is stored informally in terminologies and taxonomies - thus being not directly usable for computational reasoning on the contrary of formal ontologies - while the few existing visualization ontologies do not sufficiently represent existing domain knowledge and, furthermore, are not accessible to public. More recently, Voigt et al. (2012) have created a visualization ontology to support a recommendation system for the selection of visualization components.

## 2. METHODOLOGY

The global aim of this research is to create tools to help designers of virtual cities who must integrate the visualization of abstract data. The approach we took consists in creating a visualization selection framework that comprises an ontology of 3D visualization techniques, a knowledge base about the usability of these techniques, and selection tools that rely on this knowledge base.

**2.1 Integrated visualization selection framework**

The research methodology we used to create the ontology and the knowledge base consists in the following main phases:
- collection of the most relevant models and taxonomies of 3D visualization techniques described in the literature, then adaptation, enhancement, etc. of this knowledge in order to define a conceptualization of 3D visualization techniques for enriched 3D city models;
- formalization of those conceptualizations in OWL 2 language (OWL Working Group 2009), in order to obtain an ontology of 3D visualization techniques. This ontology represents the terminology used to describe the domain concepts using detailed axioms. The ontology editor Protégé (Protégé 2005) was used for defining the required OWL classes, properties and axioms;
- populating of the ontology from either formal or informal descriptions of 3D visualization techniques used in 3D GIS, 3D video games, etc. We thus obtain a knowledge base of visualization techniques;
- definition and formalization of the restriction rules in terms of compatibility – incompatibility between techniques. More precisely general rules of the domain or specific rules based on evaluations are used in order to define whether a combination of techniques is possible (or not), or whether a technique is relevant (or not) to be used in a certain context and /or for a specific task.

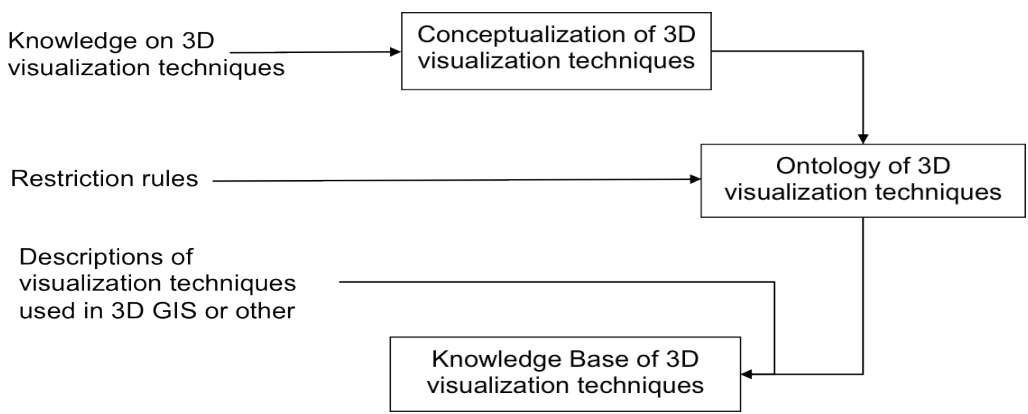

Figure 1. Methodology of the construction of an ontology of 3D visualization techniques
and of the knowledge base associated

## 2.2 Design principles for an ontology of 3D visualization techniques

Our ontology of 3D visualization techniques aims at finding the relevant visualization technique(s) (1) for a specific urban data and (2) for different data that have to be visualized within the same 3D model to perform a given task in a specific context. This is why the ontology must not only contain a collection of visualization techniques but also represent their usability relatively to a context and a task.

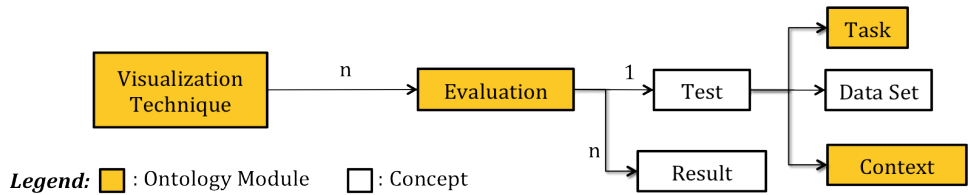

Figure 2. Specification of the interacting modules of the designed ontology for representing visualization techniques

As expressed in Figure 2, four modules compose this ontology. The first module is related to the **representation** of **visualization techniques**. The second module focuses on the **evaluation** of those techniques. The third module describes the **context** related to a specific 3D model. The fourth module is dedicated to the representation of the different tasks that a user can perform.

## 3. RESULTS AND DISCUSSION

In this contribution we focus on the module related to the representation of visualization techniques in an ontological form.

### 3.1 Ontology of visualization techniques

We describe the main concepts of our ontology of visualization techniques[1], in particular the concepts describing data and visualization techniques.

**Data concept**
A data represents the abstract information related to an object in the 3D city model and to a specific urban or environmental issue. A data is specified by:
- a **data type**, which can be, for example, a text, an audio or video element, a scalar, a vector or a combination of those elements;
- a **data format**: jpg, pdf, rdf, xml…
- the **environmental issue** to which it is related: air quality, noise…
- the **urban object**(s) to which it is related;
- a **geolocation** in the 3D model, which can be in the form of coordinates, of a geoname, object related…

A data is represented in the ontology as a concept related by properties to the previous elements defined themselves as concepts (see Figure 3).

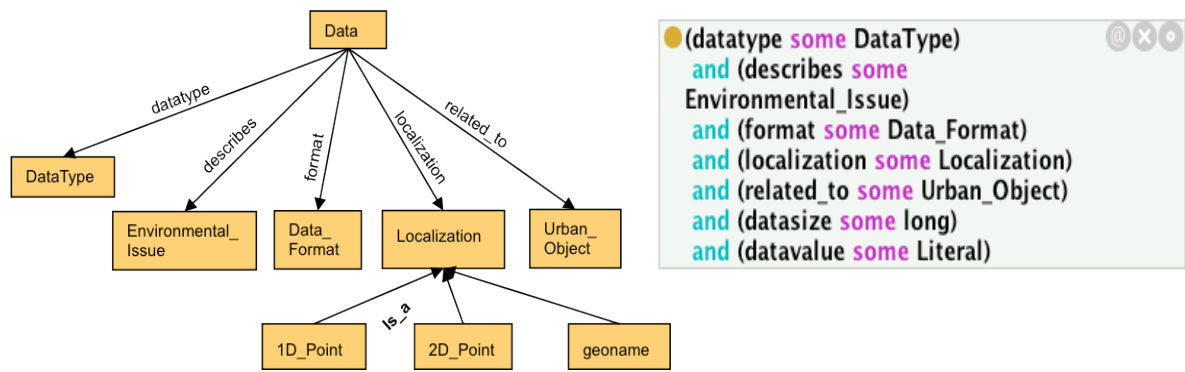

Figure 3. Specification of the Data concept (in graph form on the left, in OWL on the right)

A specific data is represented as an instance of the Data concept. Figure 4 gives two examples of the representation (in OWL) of data, a scalar value with a 3D location (on the left) and a numeric value with a 2D location (on the right):

---

[1] For more details about this ontology we invite you to view the current version published on our website: http://cui.unige.ch/isi/onto/2012/O3DVT.owl

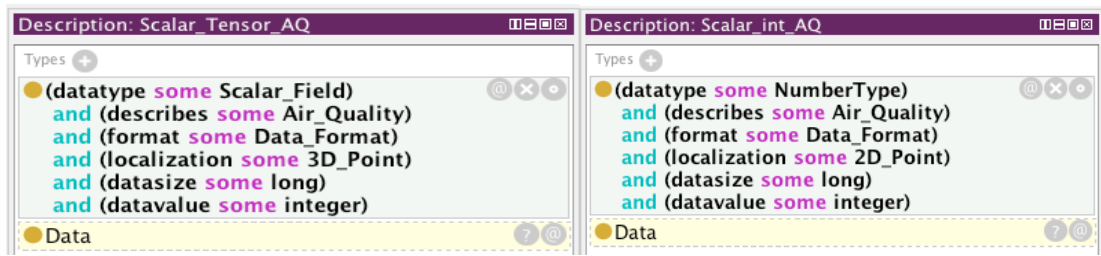

Figure 4. Examples of representation of data
a scalar value with a 3D location (on the left), a numeric value with a 2D location (on the right)

**Visualization_Technique concept**

A visualization technique is characterized by the following elements: a data, an output location, a visualization abstraction, some visual attributes, a reference and example-s (Figure 5):

- **Data**: it is the input to the technique and will be transformed into an output view based on different criteria of the layout manipulation and definition. Each data can be used as input for many visualization techniques.
- **Output Location**. The output location describes the display characteristics about the rendering of the visualization technique and the location of the view.
- **Visualization abstraction**. The visualization abstraction represents an object abstraction before rendering it on the screen. It's about information that is viewable on the screen using a visualization technique.
- **Visual attributes:** they describe the visual aspects related to the technique: whether the data support is transparent or not, whether the size is fixed or dynamic, etc.
- **Reference**: it is a link to a document describing the technique
- **Example**: it is a link to a 3D prototype or an image capture of the technique's implementation

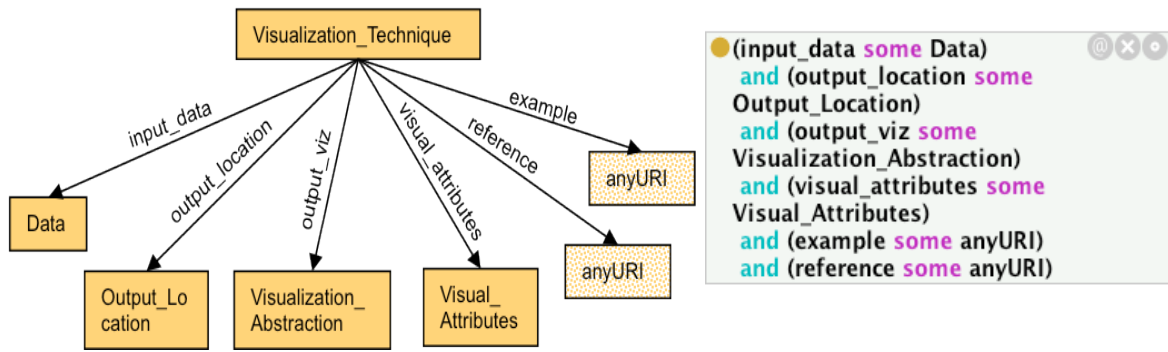

Figure 5: Specification of the Visualization_Technique concept (in graph form on the left, in OWL on the right)

A specific visualization technique is represented as an instance of the Visualization_Technique concept. Figure 6 illustrates two visualization techniques relative to air quality and issued from (San José et al 2011).

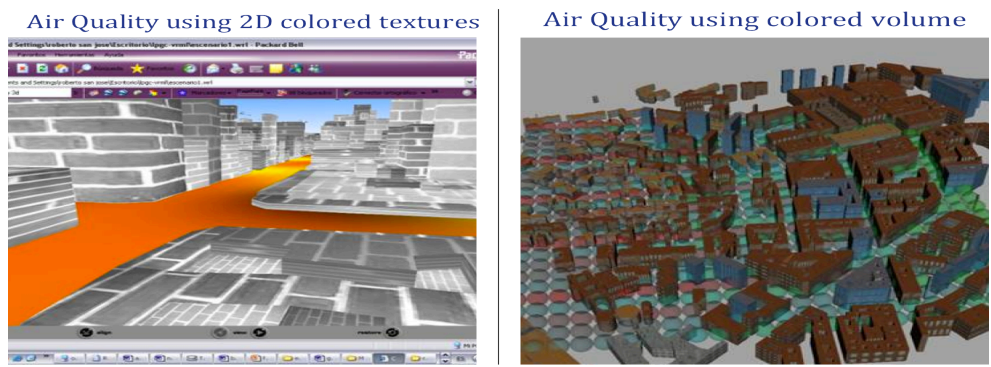

Figure 6: Two techniques (ColoredTextures on the left, ColoredBalls on the right)
for representing air quality scalar values (San José et al, 2011)

The ColoredBalls technique is named AirQuality_Scalar_VS_3D_ColoredBalls in our ontology and represented as follows:

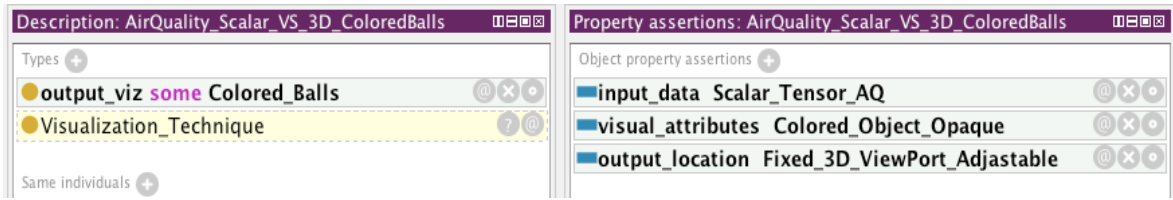

Figure 7: Representation in OWL of the visualization technique AirQuality_Scalar_VS_3D_ColoredBalls

The ColoredTextures technique is named AirQuality_Scalar_WS_2D_ColoredTextures in our ontology and represented as follows:

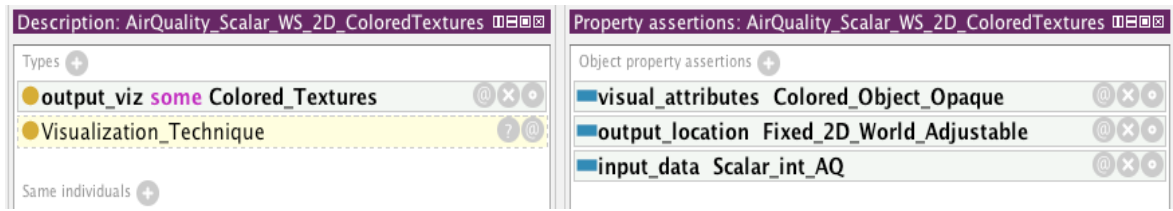

Figure 8: Representation in OWL of the visualization technique AirQuality_Scalar_WS_2D_ColoredTextures

### 3.2 Usage of the Ontology

Queries can be performed to select some relevant visualization techniques from some criteria. Figure 9 gives examples of such queries. The first query (on the left) retrieves all described visualization techniques relevant for the visualization of data in the field of air quality. The second query (on the right) retrieves all described visualization techniques (1) relevant for the visualization of numeric data in the field of air quality and (2) having an output location in a terrain as a 2D surface.

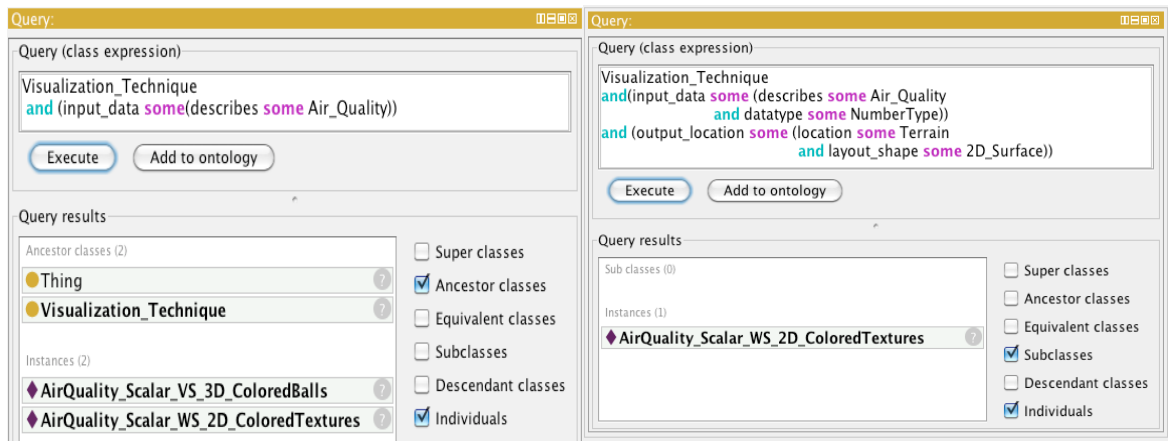

Figure 9: Querying the knowledge base to get all the techniques responding to some criteria (in OWL)

More advanced uses will involve the modules Evaluation, Task and Context. For example, in the same localization in the 3D model, we may have three data related to the same building: (1) a short text (string type) describing the building's name, (2) a set of scalar values representing air quality, and (3) a set of scalar values representing noise at the same location. Two data have the same type so they could potentially be visualized using the same technique. But the user may be confused and take the information about air quality as noise level and vice versa. An incompatibility rule can avoid such a confusion in allowing the same visualization technique being used only once. Another problem is related to possible occlusions. The representation of the label data could hide or be an obstacle for viewing the other data: thus, if the label is represented as a 3D text object it should be represented in a side or attached to the top of the building leaving enough space for representing the scalar values of noise and air quality.

## 4. CONCLUSION AND PERSPECTIVES

In this paper we have described a framework for an integrated visualization of the abstract information of enriched 3D city models. Our approach implies the construction of a 3D scene that plays the role of a "readable document" for the user. In order for such a "document" to be fully understandable and contribute to decision processes, relevant visualization techniques have to be chosen. Such a choice depends not only on the kind of abstract information or data that extends the 3D geometrical model but also on the context and the task. In order to represent the various 3D visualization techniques associated to the potentially heterogeneous data and information to display we have defined an ontology of 3D visualization techniques. Such an approach (1) enables a common description of visualization techniques that can emanate not only from the urban field but from other fields such as 3D video games, (2) provides a formalized description that can be used to select automatically the relevant techniques from the description of the data to display from a set of criteria. Restriction and composition rules will be added to the ontology in order to take in account the interaction between techniques and/or data in terms of compatibility - incompatibility.

## 5. ACKNOWLEDGEMENTS

The work described in this paper is part of the research "Designing and evaluating 3D knowledge visualization techniques for urban planning" funded by the swiss "Secrétariat d'état à la recherche" (SER No C10.0150) in relation with the COST Action TU0801 "Semantic enrichment of 3D city models for sustainable urban development".